# CarbonFish: A Bistable Compliant Fish Robot Capable of High-Frequency Undulation*

Zechen Xiong[1, *], Zihan Guo[2], Mark Liu[2], Jialong Ning[2], Hod Lipson[2]

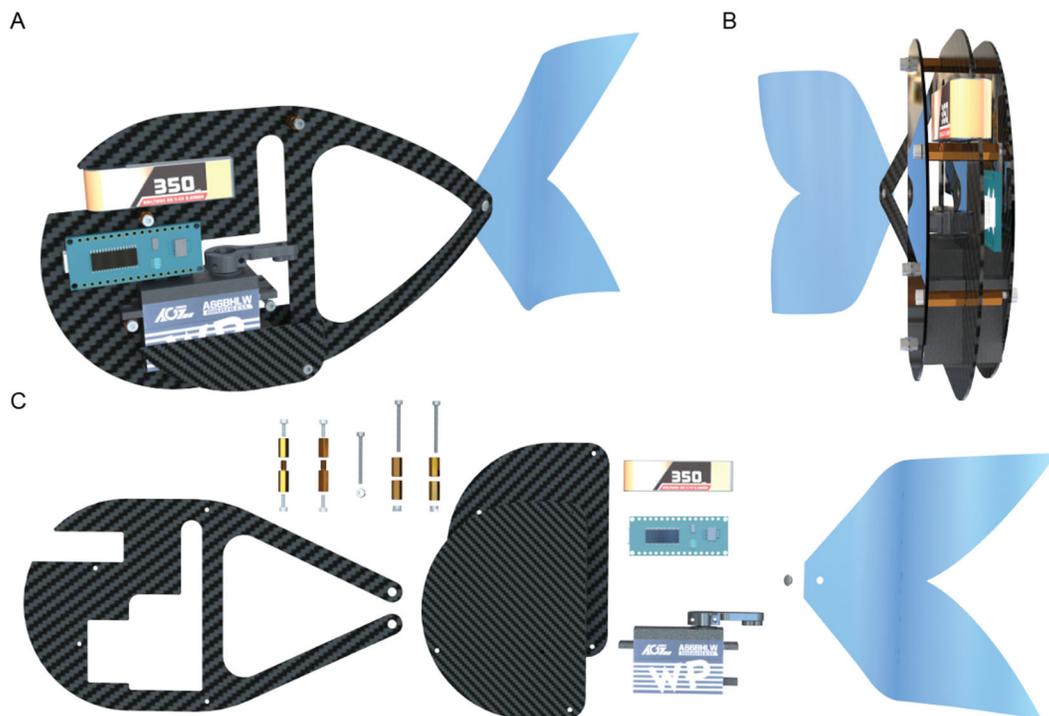

Figure 1 Hair-clip mechanism (HCM) compliant fish robots. (A) The side view of CarbonFish. Components include 0.5mm CFRP plates, PETG plastic film, onboard 7.6V Li-po battery pack, Arduino Nano 33, A66BHLW waterproof servo motor, and a variety of mechanical fasters (screws, nuts, spacers, non-slip nuts, etc.). (B) The front view of CarbonFish. (C) The exploded view of CarbonFish.

*Abstract*—The Hair Clip Mechanism (HCM) is a kind of innovatively structured bistable mechanism made from 2D materials, as described in preceding studies [1], [2], [3], [4]. It can be used to increase the functionality of soft robotics, like soft robotic fish, crawlers, and grippers. Compared with unstructured soft robotic systems, HCM robotics have the advantage of augmented mobility and design simplicity. In this research, we investigate the paradigm of the design and fabrication of an HCM-based fish robot that is capable of high-frequency undulation. A fish robot prototype made of carbon fiber-reinforced plastic (CFRP) 2D ribbon is proposed and assembled based on the theory and design paradigm herein referred to as "CarbonFish." Preliminary evaluations of our single-actuated CarbonFish have evidenced an undulation frequency approaching 10 Hz, suggesting its potential to achieve high-speed and biomimetic swimming.

## I. INTRODUCTION

Soft and compliant robotics represents an advancing domain in robotics research, emphasizing the design and development of robots utilizing soft and deformable materials. The intrinsic flexibility of these materials facilitates robots to replicate biomechanical movements, allowing for adaptive interactions with their environment [5], [6]. Specifically, soft robotic fish have obtained significant attention, given their prospective applications in non-intrusive underwater exploration and systematic environmental monitoring [7]. Contrary to traditional propeller-driven underwater vehicles, these biomimetic robots usually employ fluidic or electroactive polymer actuators, propelling themselves via repeated undulations. Several studies underscore the potential of this technology. For instance, Marchese et al., 2012 [8] detail the design, fabrication, and

*Research supported by the Fu Foundation School of Engineering and Applied Science, Columbia University and U.S. National Science Foundation (NSF) AI Institute for Dynamical Systems grant 2112085.

[1]Zechen Xiong is with the Dept. of Earth and Environment Engineering at Columbia University, New York, NY 10027 USA (phone: 9173023864, e-mail: zechen.xiong@columbia.edu).

[2]Zihan Guo, Mark Liu, Jialong Ning, and Hod Lipson are with the Dept. of Mechanical Engineering at Columbia University, New York, NY 10027 USA (e-mail: zg2450@columbia.edu, ml2877@columbia.edu, jn2894@columbia.edu, and hod.lipson@columbia.edu).

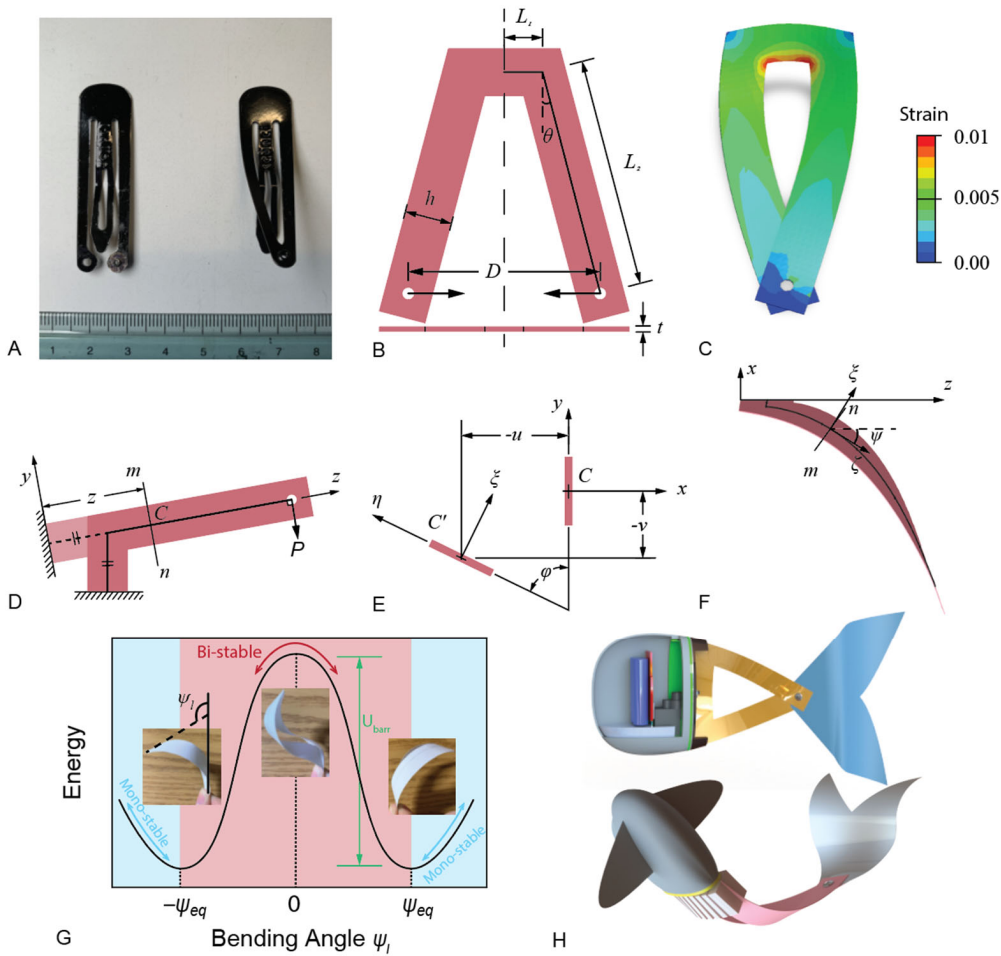

Figure 2 The principle, modeling, and prototypes of HCM soft robotics. Adapted from Xiong et al. [2]. (A) A steel hair clip before and after assembly. (B) and (C) The characterization and assembly of a typical HCM. (D) - (F) The mathematical modeling of the hair clip mechanism (HCM). Coordinate $z$ is defined as the straightened coordinate of the half ribbon. (G) The energy landscape of a paper HCM. (H) HCM fish robots from the previous studies.

experimental verification of a complete, tetherless, pressure-operated soft robotic platform. Katzschmann et al. [9] present an autonomous, soft-bodied robotic fish that is hydraulically actuated and capable of sustained swimming in three dimensions. Marchese et al. [10] describe an autonomous soft-bodied robot that is both self-contained and capable of rapid, continuum-body motion. Katzschmann et al. [7] have elucidated the design, fabrication, control mechanisms, and marine evaluations of a particular soft robotic platform. This fish robot exhibits a lifelike undulating tail motion enabled by a soft robotic actuator design that can potentially facilitate a more natural integration into the ocean environment, demonstrating controlled navigation in the natural aquatic environments and proficiency in conducting detailed marine ecological assessments. Berg et al. [11] made OpenFish. A detailed description of the design, construction, and customization of the soft robotic fish is presented. Clapham and Hu [12], [13], [14], [15] developed a series of motor-driven tethered/untethered fish robots, iSplash-I, iSplash-MICRO, iSplash-II, and iSplash-OPTIMIZE, that range from 50-620 mm in length and that can operate at 5-20 Hz, providing a lab-condition speed of 3-11.6 BL/s (up to 3.7 m/s) that is comparable to real fish locomotion speed. Other influential works include Li et al. [16], Li et al. [17], and Lin et al. [18].

While soft robotic fish show great promise, several challenges remain. First of all, the highest speed achieved by this type of fish robot is only about 0.5 [7] ~ 0.7 BL/s [16]. Others may have demonstrated faster speed but are not adequate in a tetherless situation [11], [16]. Second, the performance of such robots is significantly influenced by their empirical design and manual fabrication. We propose to use hair clip mechanisms (HCMs), an energy-storing-and-releasing mechanism, to address these problems. Hair clip mechanism (HCM) is an in-plane prestressed bistable mechanism proposed in our previous research [1], [2], [3], [4], [19] to enhance the functionality of soft robotics. HCMs have several advantages, such as high mobility, good repeatability, and design- and fabrication- simplicity. This work delves into designing a novel HCM robotic propulsion system made from PETG plastic, carbon fiber-reinforced plastic (CFRP), and steel. Specifically, a CFRP HCM fish robot, which we term CarbonFish (Figure 1), is made and can operate at frequencies up-to-10 Hz. Detailed derivation and verification of the HCM theory are given, and the influence of key parameters like dimensions, material types, and servo motor specifications are summarized. The designing algorithm offers insight into HCM robotics. It enables us to search for suitable components, operate robots at a desired frequency, and achieve high-frequency and high-speed undulatory swimming for fish robots.

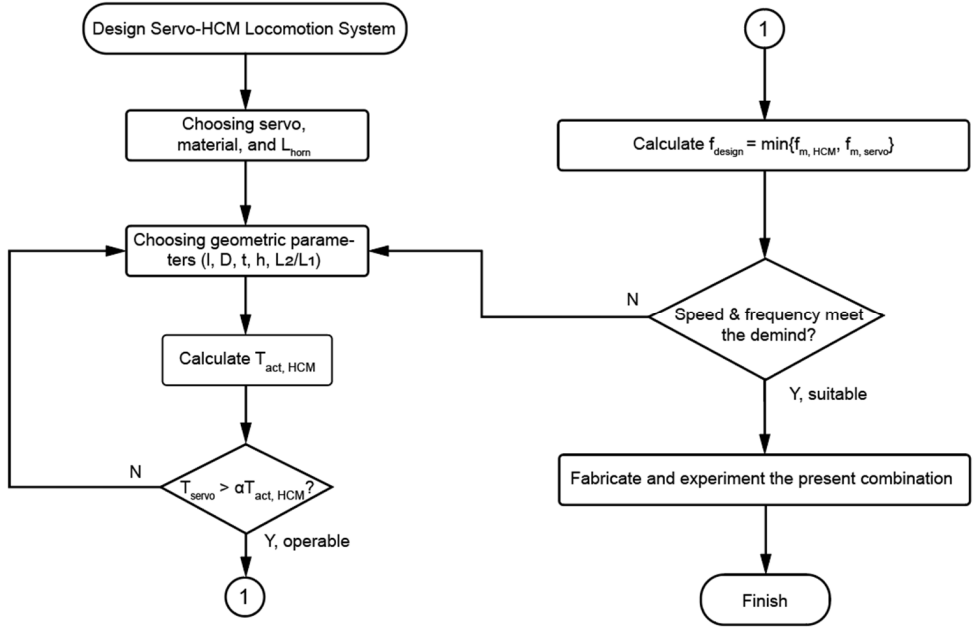

Figure 3 Schematic showing the designing algorithm of servo-HCM robotic systems based on Eq. (9)-(11). The design factor $\alpha$ is assumed to be 1.0 in our cases.

## II. WORKING PRINCIPLES

The inspiration, assembly, mathematical modeling, and robotic applications of HCMs are depicted in [2] and Fig. 2. The torsional angular displacement $\varphi$ (Figure 2E) of a typical HCM can be depicted by the equation

$$\varphi(z) = \sqrt{l-z}\, A_1 J_{1/4}\left(2.7809\left(\frac{l-z}{l}\right)^2\right). \quad (1)$$

in which $l = L_1 + L_2$ is the half ribbon length, $A_1$ is a non-zero integration constant that can be determined by considering energy conservation law, and $J_{1/4}$ is the Bessel function of the first kind of order. The translational displacement

$$u(z) = \int_0^z \varphi(z)\,\mathrm{d}z, \quad (2)$$

Please note that $z$'s in the integrand $\varphi(z)\,\mathrm{d}z$ are dummy variables that cancel out upon integration.

The energy barrier between the bi-states of the HCM can be approximated as [2]

$$U_{\text{barr}} = 3 P_{cr} \cdot D, \quad (3)$$

in which $D$ is the prestressing displacement in Figure 2B. Assuming Hooke's law (linear elasticity), when the servo deforms HCM, the maximum torque required to snap the HCM can be calculated as

$$T_{\text{act, HCM}} = 2 U_{\text{barr}} \cdot L_{\text{horn}} / 2u(L_1) \quad (4)$$

in which $L_{\text{horn}}$ is the length of the servo horn, and $L_1$ is the core section length of HCM in Figure 2B. The torque capacity of the servo is taken as the stall torque. The frequency capacity of the servo-HCM system can be calculated as

$$f_{\text{design}} = \min\begin{cases} f_{m,\text{ HCM}} = 1/2 t_*, \\ f_{m,\text{ servo}} = speed / 4u(L_1) \end{cases}, \quad (5)$$

in which $t^*$ is the timescale of the HCM snapping and is estimated as[20]

$$t_* = \frac{(2l)^2}{t\sqrt{E/\rho_s}}. \quad (6)$$

in which $\rho_s$ denotes the material density.

## III. RESULTS

### A. Servo-HCM designing

According to the HCM working principles, the algorithm for designing a servo-HCM locomotion system fish robots can be illustrated in Figure 3. Usually, the most important and time-consuming procedure is to look for correct combinations of servo, material, and shapes, according to our experience with the two fish robots. Typical materials include plastic, CFRP, and steel sheets, whose related properties are given in Table I. These materials have high Young's modulus, tensile strength, and elastic strain limits. When they are in the form of sheets, their out-of-plane stiffness is usually smaller than their in-plane stiffness (membrane stiffness) by an order of $10^5 \sim 10^6$, making these 2D structures bendable and compliant. The information on common servo motors is provided in Table II, in which $T_{\text{servo}}$ denotes the stall torque of servo motors. A design factor of $\alpha > 1$ is suggested to ensure the successful servo-driven snap-through buckling of the HCM. Since the undulating frequency capacity is the most important parameter for a fish robot to achieve high speed, the present algorithm is used to iterate the geometry for a satisfactory design.

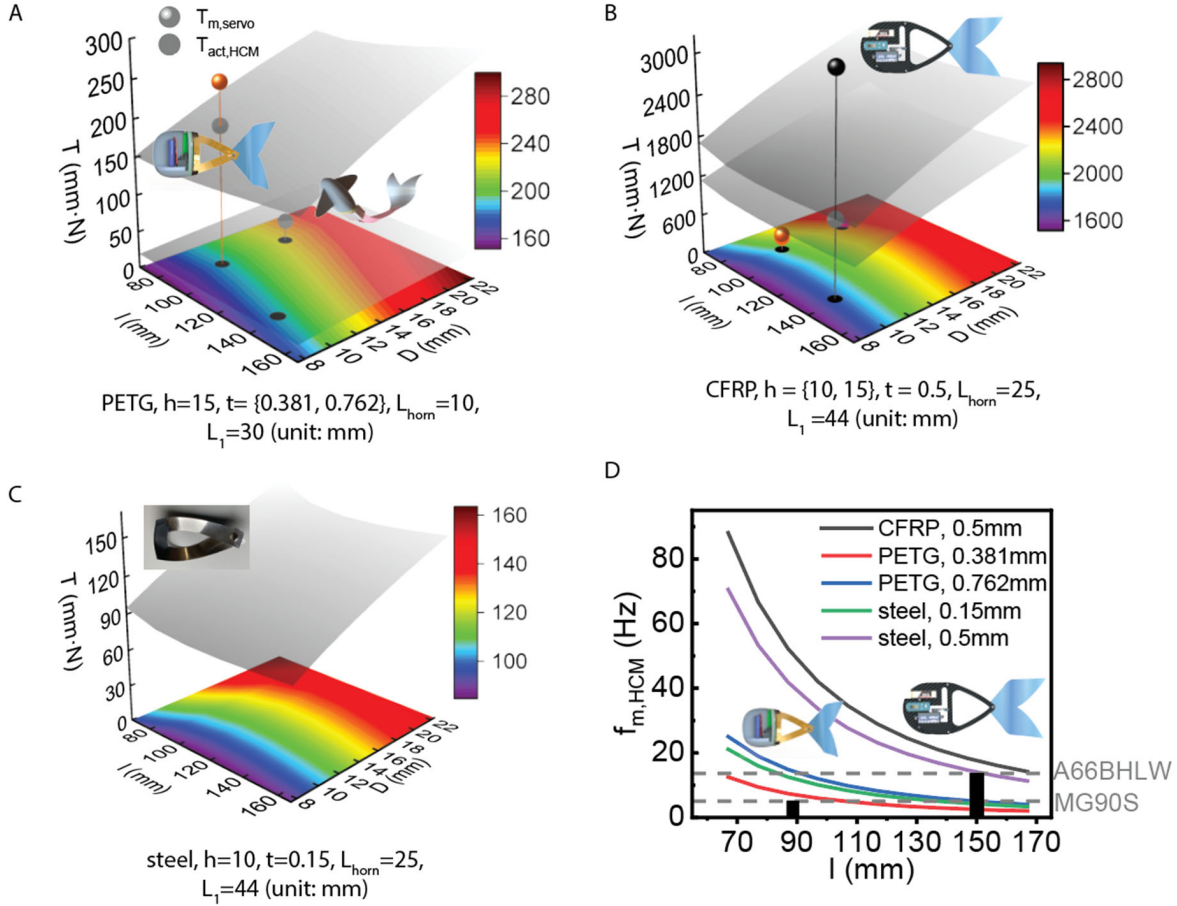

Figure 4. The influence of materials and dimensions on the actuation torque $T_{act, HCM}$, and the design frequency $f_{design}$. (A)-(C) The $T_{act, HCM}$ profile w.r.t geometric parameters ($l, D, t, h, L_2/L_1$) for materials of PETG plastic sheets, CFRP sheets, and steel sheets, respectively. The $T_{act, HCM}$ is found to be proportional to $D$, $t^3$, and $h$ but is less affected by $l$. Actual prototypes are embedded in the surface plots for comparison and validation. (D) The design frequency capacity $f_{design}$ calculation for different materials, geometry, and servos. Both the coral fish and CarbonFish are limited by the speed of the servos.

TABLE I. DIMENSIONS AND MECHANICAL PROPERTIES OF TYPICAL HCM MATERIALS

| Material | $t$ (mm) | $\rho_s$ (t/mm³) | $E$ (MPa) | $E/\rho_s$ (mJ/t) |
|---|---|---|---|---|
| PETG | 0.381, 0.762 | 1.25e-9 | 1.7e3 | 1.42e12 |
| CFRP | 0.5, 0.79 | 1.6e-9 | 64e3 | 40e12 |
| steel | 0.15, 0.5 | 7.8e-9 | 200e3 | 25e12 |

TABLE II. SPECIFICATIONS OF COMMON SERVO MOTORS

|  | $T_{servo}$ (mm·N) | speed (rad/s) | weight (g) | $L_{horn}$ (mm) | $f_{m, servo}$ (Hz) |
|---|---|---|---|---|---|
| MG90S | 245 | 10.5 | 14 | 10 | 4.5 |
| B24CLM | 588 | 12.3 | 22 | 20 | 6.15 |
| A66BHLW | 3234 | 15.4 | 66 | 25 | 13.6 |
| A06CLS | 294 | 20.1 | 7 | 13 | 17.0 |
| DS3230MG | 3381 | 6.16 | 58 | 25 | 3.08 |
| SG92R | 245 | 10.5 | 9 | 10 | 4.5 |
| ZOSKAY | 3430 | 9.5 | 60 | 25 | 4.76 |

Based on the HCM theory, the influence of geometric parameters is illustrated in Figure 4. Besides the material type and servo horn, dimensions ($l, D, t, h, L_2/L_1$) decide the required actuation torque $T_{act, HCM}$ of HCMs. More specifically, we find $T_{act, HCM}$ is proportional to $D$, $h$, and $t^3$, and is less relevant to $l$ since the increase of $l$ not only decreases energy barrier $U_{barr}$ but also decreases $2u(L_1)$, i.e., the required actuation displacement of HCM snap-through. Figure 4, with the three prototypes embedded, is to validate the fabrication methodology and delineate the influence of geometric parameters. The pink fish is made from PETG HCM that has a geometry of ($l, D, t, h, L_2/L_1$) = (87.5, 17.1, 0.381, 15, 6) (unit: mm and 1), which corresponds to a required torque of $T_{act, HCM}$ = 28.3 mm • N if using a servo motor to actuate; the coral fish robot has a shape of ($l, D, t, h, L_2/L_1$) = (87, 11.8, 0.762, 15, 1.9), corresponding to $T_{act, HCM}$ = 188.7 mm • N and thus a design factor $\alpha_{coral}$ = 245/188.7 = 1.3 due to the use of an MG90S. Steel sheets have not been used yet but should have a similar effect on $T_{act, HCM}$ as plotted in Figure 4C when used with a servo horn of 25 mm (A66BHLW).

The frequency capacity analysis is demonstrated in Figure 4D. According to Eq. (11), the $f_{m, HCM}$ is proportional to $t$ and the reciprocal of $l^2$. However, the limiting condition is usually the speed of the servos, as in Figure 4D. For example, the coral fish has a maximum HCM frequency of $f_{m, HCM}$ = 14.8 Hz, while the MG90S servo operates at a maximum of $T_{m,servo}$ = 4.5 Hz. Its actual achievable undulation frequency is 0 ~ 4 Hz[2].

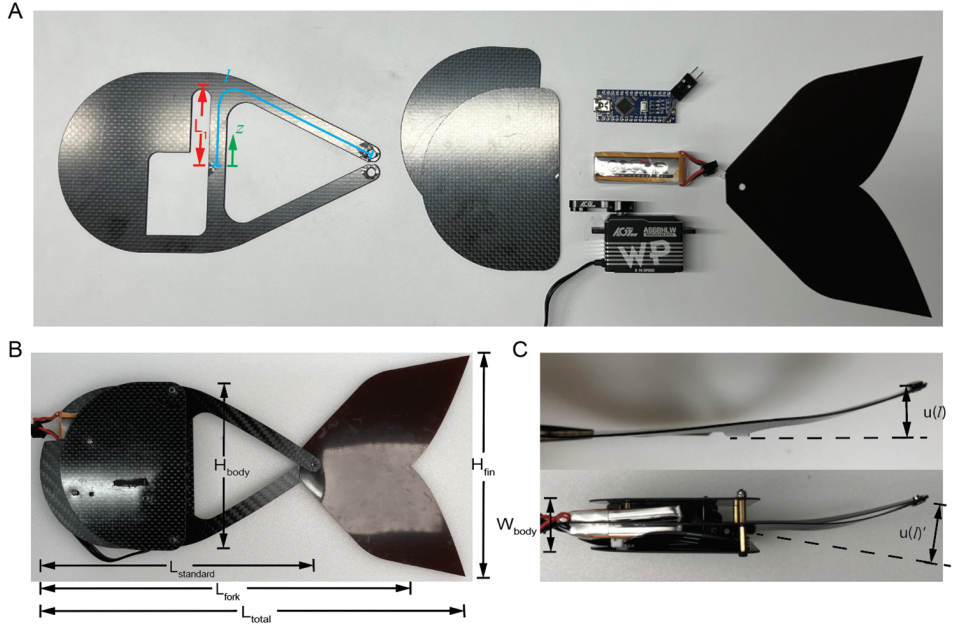

Figure 5. Photos of CarbonFish. (A) Individual components of CarbonFish. (B) Side view and dimensions of CarbonFish. (C) Bottom view and dimensions of CarbonFish. Parameters $H_{body}$ = 100 mm, $H_{fin}$ = 140 mm, $L_{standard}$ = 170 mm, $L_{fork}$ = 232 mm, $L_{total}$ = 270 mm, $W_{body}$ = 21 mm, and $u(l)$ = 25 mm ≈ $u'(l)$.

*B. Fabrication*

The HCM theory shows that the matrix material of HCMs plays a significant role in its performance. The timescale and the energy barrier are highly determined by Young's modulus of the material. Meanwhile, the repetitive snap-through buckling requires the material to be elastic and strong, which corresponds to the elastic limit and the strength of the material. CFRP sheets are well-known for their high modulus, yield strain, tensile strength, and light-weightedness[21], and they are very promising to be an excellent component for HCM-compliant robots. From TABLE I, CFRP has the highest specific Young's modulus $E/\rho_s$, which indicates its potential to generate a high-frequency capacity $f_{design}$. Also, the thin thickness of CFRP can guarantee a low enough $T_{act, HCM}$ for the successful snapping of HCMs. Thus, we fabricate CarbonFish with ($l$, $D$, $t$, $h$, $L_2/L_1$) = (137, 10, 0.5, 10, 2.1), which gives a design factor $\alpha_{carbon}$ = 3234/1177 = 2.8 with A66BHLW as the driving servo (Figure 4B and 5). The components of CarbonFish are shown in photos of Figure 5, with parameters body height $H_{body}$ = 100 mm, fin height $H_{fin}$ = 140 mm, Standard length $L_{standard}$ = 170 mm, fork length $L_{fork}$ = 232 mm, total length $L_{total}$ = 270 mm, and body width $W_{body}$ = 21 mm. Assuming the lateral displacement before the assembly of the fish $u(l)$ equals the lateral displacement $u'(l)$ afterward, their theoretical value is 36 mm from Eq. (7), and actual values are $u(l)$ = 25 mm ≈ $u'(l)$. The deviation of the experiment from theory is due to the strengthening of the core area's (denoted by $L_1$) bending stiffness by the body plates.

Limited by the speed of the servo used, CarbonFish has a theoretic frequency capacity of $f_{design} = f_{m, servo}$ = 13.6 Hz, and can achieve 10 Hz undulation with the prototype (movie S1[1]). To further increase the flapping frequency and swimming speed in the future, a DC motor-driven HCM system would be very beneficial due to the DC motor's high rotating speed.

The hydraulic friction of swimmers mainly consists of two parts: the resistance created by the pressure difference between the front and rear water, which is approximately proportional to the sectional area of the swimmer, and the resistance created by the boundary layer, which is approximately proportional to the surface area of the swimmer. In CarbonFish, most components are open to water flow to reduce the resistance from the sectional area. However, the waterproofing of these electronics would be a big challenge.

Figure 6 compares our inventions with the existent fish robots, soft or rigid. Although the speed of HCM fish robots is not as high as the state-of-the-art ones[22], their velocities in body length per beat are very promising, ranging from 0.34 ~ 0.54 BL/beat. Based on that, the 10-Hz-undulating CarbonFish is estimated to have a speed of 6.8 ~ 10.8 BL/s, as shown in Figure 6A, and has the potential to compete with real fish that swim in the range of 2 ~ 10 BL/s[23].

IV. CONCLUSION

The research depicts the development and validation of the Hair Clip Mechanism (HCM) for application in soft robotic fish, specifically the CarbonFish model. The HCM, an in-plane prestressed bistable mechanism, has been shown to significantly enhance the structural rigidity and functional mobility of soft robotics compared to other soft robotic designs. Besides, the HCM's bistable feature allows for energy storage and release, which is harnessed to produce a novel undulation pattern that combines energy efficiency with high propulsion. The HCM's ability to function as a load-bearing skeleton, motion transmission system, and high-speed actuator is a testament to the versatility and potential of this mechanism in soft robotics.

---
[1] https://youtu.be/KkzBR8yb-OY

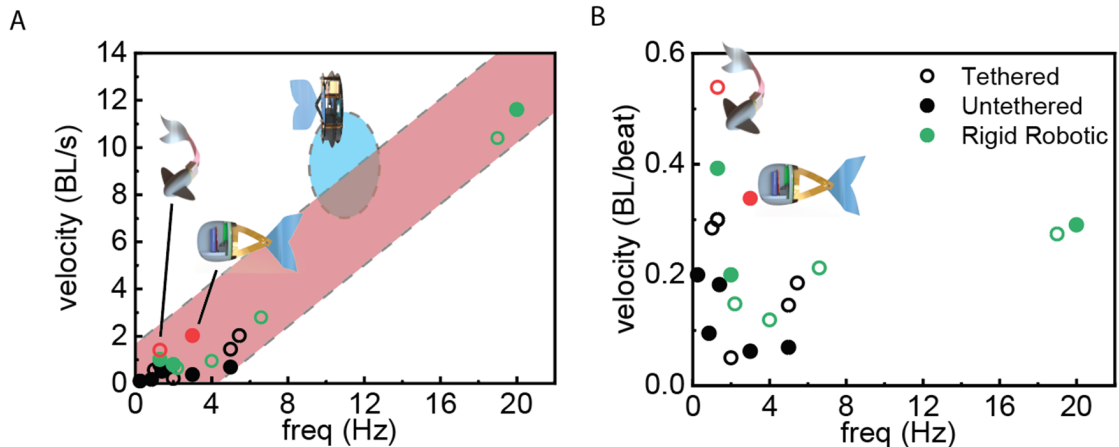

Figure 6 The comparison between HCM fish robots and existent fish robots from literature[8], [12], [13], [17], [18], [21], [23], [25], [28], [32], [33], [34], [35], [36], [37], [38]. (A) The speed comparison in body length per second (BL/s). (B) The speed comparison in body length per beat (BL/beat).

The utilization of carbon fiber-reinforced plastic (CFRP) as the core material for the HCM has been a pivotal aspect of this study, contributing to the high-frequency undulatory capabilities of CarbonFish. The CarbonFish, with its single-actuated design, has demonstrated undulation frequencies of up to 10 Hz, indicating a potential to achieve swimming speeds that could rival or surpass those of real fish. The design and fabrication methodology, underpinned by mathematical modeling, has been critical in achieving these results. Future work may involve exploring alternative actuation systems, such as DC motor-driven HCM systems, to further enhance the undulation frequency and swimming speed. Additionally, addressing the waterproofing challenges presented by the open component design will be crucial for practical applications.